\documentclass{article}

\usepackage{times}
\usepackage{graphicx} 
\usepackage{subfigure} 

\usepackage{natbib}

\usepackage{algorithm}
\usepackage{algorithmic}

\usepackage{hyperref}

\usepackage[accepted]{icml2017}

\icmltitlerunning{Handwritten Bangla Digit Recognition Using Deep Learning}

\begin{document} 

\twocolumn[
\icmltitle{Handwritten Bangla Digit Recognition Using Deep Learning}
\icmlauthor{Md Zahangir Alom}{alomm1@udayton.edu}
\icmladdress{University of Dayton, Dayton, OH, USA}
\icmlauthor{Paheding Sidike}{pahedings1@udayton.edu}
\icmladdress{University of Dayton, Dayton, OH, USA}
\icmlauthor{Tarek M. Taha}{ttaha1@udayton.edu}
\icmladdress{University of Dayton, Dayton, OH, USA}
\icmlauthor{Vijayan K. Asari}{vasari1@udayton.ed}
\icmladdress{University of Dayton, Dayton, OH, USA}

\icmlkeywords{deep learning, inception network, object recognition}
\vskip 0.3in
]

\begin{abstract}
	In spite of the advances in pattern recognition technology, Handwritten Bangla Character Recognition (HBCR) (such as alpha-numeric and special characters) remains largely unsolved due to the presence of many perplexing characters and excessive cursive in Bangla handwriting. Even the best existing recognizers do not lead to satisfactory performance for practical applications. To improve the performance of Handwritten Bangla Digit Recognition (HBDR), we herein present a new approach based on deep neural networks which have recently shown excellent performance in many pattern recognition and machine learning applications, but has not been throughly attempted for HBDR. We introduce Bangla digit recognition techniques based on Deep Belief Network (DBN), Convolutional Neural Networks (CNN), CNN with dropout, CNN with dropout and Gaussian filters, and CNN with dropout and Gabor filters. These networks have the advantage of extracting and using feature information, improving the recognition of two dimensional shapes with a high degree of invariance to translation, scaling and other pattern distortions. We systematically evaluated the performance of our method on publicly available Bangla numeral image database named CMATERdb $3.1.1$. From experiments, we achieved $98.78\%$ recognition rate using the proposed method: CNN with Gabor features and dropout, which outperforms the state-of-the-art algorithms for HDBR. 
	
	
\end{abstract}

\section{Introduction}
\label{intro}
Automatic handwriting character recognition is of academic and commercial interests. Current algorithms are already excel in learning to recognize handwritten characters. The main challenge in handwritten character classification is to deal with the enormous variety of handwriting styles by different writers in different languages. Furthermore, some of the complex handwriting scripts comprise different styles for writing words. Depending on languages, characters are written isolated from each other in some cases, (e.g., Thai, Laos and Japanese). In some other cases, they are cursive and sometimes the characters are connected with each other (e.g., English, Bangladeshi and Arabic). These challenges are already recognized by many researchers in the field of Natural Language Processing (NLP) \cite{1,2,3}. Handwritten character recognition is more difficult comparing to printed forms of characters. This is because characters written by different people are not identical and varies in different aspects such as size and shape. Numerous variations in writing styles of individual characters also make the recognition task challenging. The similarities in different character shapes, the overlaps, and the interconnections of the neighboring characters further complicate the character recognition problem.  In other words, the large variety of writing styles, writers, and the complex features of handwritten characters are very challenging for accurately classifying the hand written characters. 

\begin{figure*}
	\centering
	\includegraphics[scale=0.6]{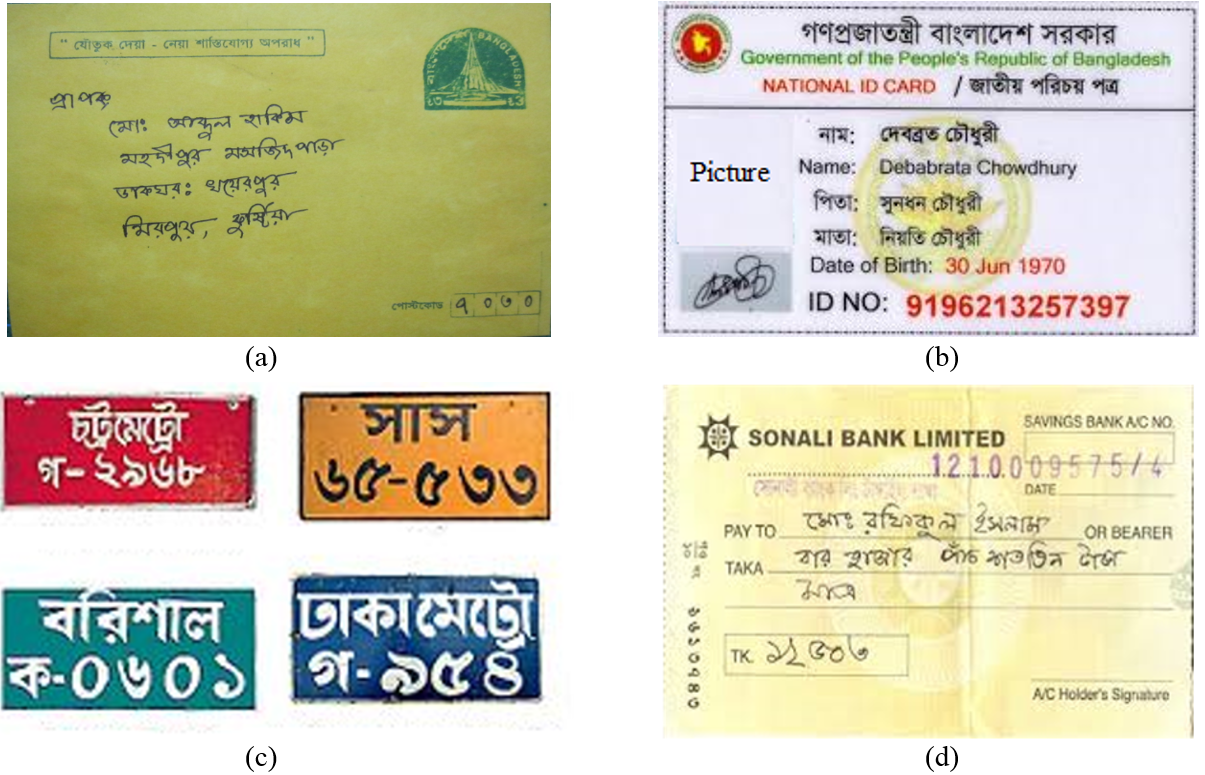}
	\caption{Example images of Banagla digits in real-life: (a)Envelope digits, (b)national ID card, (c) license plate, and (d)Bank check.}
	\label{fig:fig-1}       
\end{figure*}

Bangla is one of the most spoken languages, ranked fifth in the world. It is also a significant language with a rich heritage; February 21st is announced as the International Mother Language day by UNESCO to respect the language martyrs for the language in Bangladesh in 1952. Bangla is the first language of Bangladesh and the second most popular language in India. About 220 million people use Bangla as their speaking and writing purpose in their daily life. Therefore, automatic recognition of Bangla characters has a great significance. Different languages have different alphabets or scripts, and hence present different challenges for automatic character recognition with respect to language. For instance, Bangla uses a Sanskrit based script which is fundamentally different from English or a Latin based script. The accuracy of character recognition algorithms may vary significantly depending on the script. Therefore, Handwritten Bangla Character Recognition (HBCR) methods should be investigated with due importance. There are 10 digits and 50 characters in vowel and consonant in Bangla language where some contains additional sign up and/or below. Moreover, Bangla consists with many similar shaped characters; in some cases a character differ from its similar one with a single dot or mark. Furthermore, Bangla language also contains with some special characters in some special cases. That makes difficult to achieve a better performance with simple technique as well as hinders to the development of HBCR system. In this work, we investigate HBCR on Bangla digits. There are many application of Bangla digit recognition such as: Bangla OCR, National ID number recognition system, automatic license plate recognition system for vehicle, parking lot management, post office automation, online banking and many more. Some example images are shown in Fig. \ref{fig:fig-1}. Our main contributions in this paper are summarized as follows:

\begin{itemize}
	\item To best our knowledge, this is the first research conducted on Handwritten Bangla Digit Recognition (HBDR) using Deep Learning(DL) approaches.
	\item An integration of CNN with Gabor filters and Drop-out is proposed for HBDR.
	\item A comprehensive comparison of five different DL approaches are presented.
\end{itemize}

\section{Related works}
There are a few remarkable works available for HBCR. Some literatures have reported on Bangla numeral recognition in past few years \cite{4,5,6}, but there is few research on HBDR who reach to the desired result. Pal et al. have conducted some exploring works for the issue of recognizing handwritten Bangla numerals \cite{9,7,8}. Their proposed schemes are mainly based on the extracted features from a concept called water reservoir. Reservoir is obtained by considering accumulation of water poured from the top or from the bottom of numerals. They deployed a system towards Indian postal automation. The achieved accuracies of the handwritten Bangla and English numeral classifier are $94\%$ and $93\%$, respectively. However, they did not mention about the recognition reliability and the response time in their works, which are very important evaluation factors for a practical automatic letter sorting machine. Reliability indicates the relationship between error rate and recognition rate. Liu and Suen \cite{10} showed the recognition rate of handwritten Bangla digits on a standard dataset, namely the ISI database of handwritten Bangla numerals \cite{11}, with 19392 training samples and 4000 test samples for 10 classes (i.e., 0 to 9) is $99.4\%$. Such high accuracy has been attributed to the extracted features based on gradient direction, and some advanced normalization techniques. Surinta et al. \cite{12} proposed a system using a set of features such as the contour of the handwritten image computed using 8-directional codes, distance calculated between hotspots and black pixels, and the intensity of pixel space of small blocks. Each of these features is used for a nonlinear Support Vector Machine (SVM) classifier separately, and the final decision is based on majority voting. The data set used in \cite{12} composes of 10920 examples, and the method achieves an accuracy of $96.8\%$. Xu et al. \cite{13} developed a hierarchical Bayesian network which takes the database images directly as the network input, and classifies them using a bottom-up approach. An average recognition accuracy of $87.5\%$ is achieved with a data set consisting 2000 handwritten sample images. Sparse representation classifier for Bangla digit recognition is introduced in \cite{14}, where the recognition rate of $94\%$ was achieved. In \cite{15}, the basic and compound character of handwritten Bangla recognition using Multilayer Perception (MLP) and SVM classifier are achieved around $79.73\%$ and $80.9\%$ accuracy, respectively. HBDR using MLP was presented in \cite{16} where the average recognition rate using 65 hidden neurons reaches $96.67\%$. Das et al. \cite{17} proposed a genetic algorithm based region sampling strategy to alleviate regions of the digit patterns that having insignificant contribution on the recognition performance. Very recently, Convolutional Neural Network (CNN) is employed for HBCR \cite{18} without any feature extraction in priori. The experimental results shows that CNN outperforms the alternative methods such as hierarchical approach. However, the performance of CNN on HBDR is not reported in their work.

\section{Proposed scheme}
\subsection{Deep learning}
In the last decade, deep leaning has proved its outstanding performance in the field of machine learning and pattern recognition. Deep Neural Networks (DNN) generally include Deep Belief Network (DBN), Stacked Auto-Encoder (SAE) and CNN. Due to the composition of many layer, DNNs are more capable for representing the highly varying nonlinear function compared to shallow learning approaches \cite{19}. Moreover, DNNs are more efficient for learning because of the combination of feature extraction and classification layers. Most of the deep learning techniques do not require feature extraction and take raw images as inputs followed by image normalization. The low and middle levels of DNNs abstract the feature from the input image whereas the high level performs classification operation on the extracted features.The final layer of DNN uses a feed-forward neural network approach. As a result, it is structured as a uniform framework integrated with all necessary modules within a single network. Therefore, this network model often lead to better accuracy comparing with training of each module independently. 

According to the structure of the Multilayer Backpropagation (BP) algorithm, the error signal of the final classification layer is propagated through layer by layer to backward direction while the connection weights are being updated based on the error of the output layer. If the number of hidden layers becomes large enough, the BP algorithm performs poorly which is called “diminishing gradient problem”. This problem happens because the error signal becomes smaller and smaller, and it eventually becomes too small to update weights in the first few layers. This is the main difficulty during the training of NNs approach. 

However, Hinton et al. \cite{20} proposed a new algorithm based on greedy layer-wise training to overcome the “diminishing gradient problem” which leads to DBN. In this approach, first pre-training the  weights using unsupervised training approach from the bottommost layer. Then, fine-tune the weights using supervised approach to minimize the classification errors \cite{21}. This work made a breakthrough that encouraged deep learning research. Moreover, the unsupervised part is updated using another neural network approach called Restricted Boltzmann Machine (RBM)\cite{22}. 

\begin{figure*}
	\includegraphics[scale=0.705]{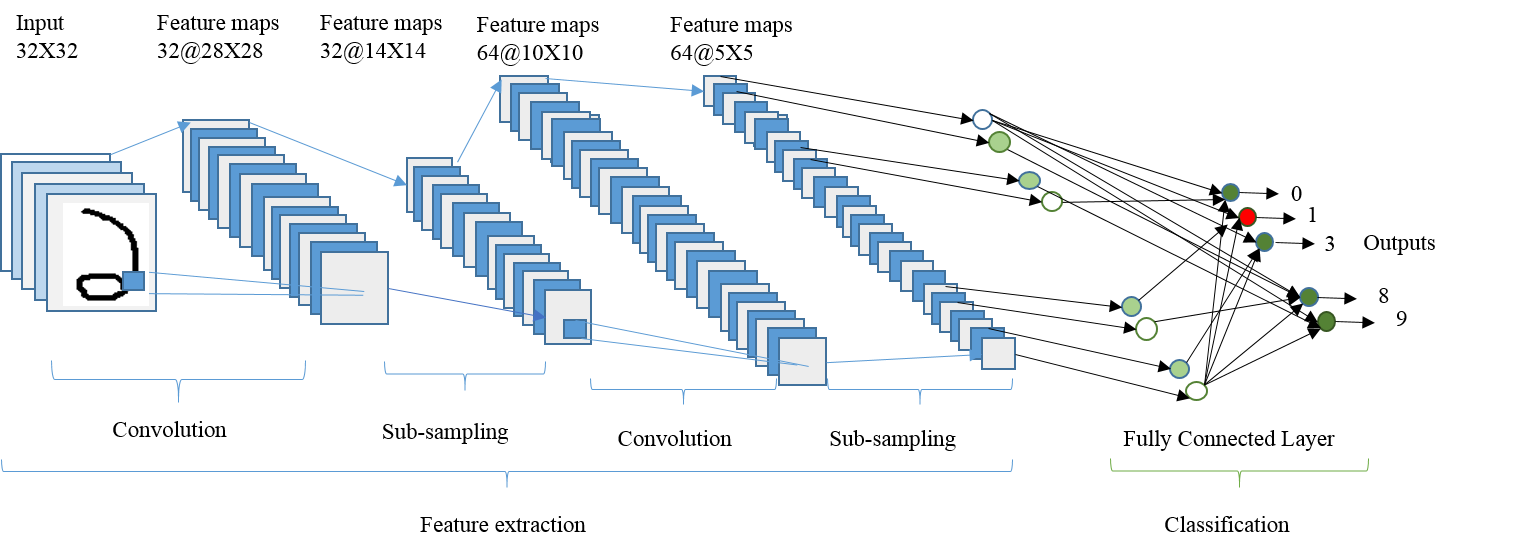}
	\caption{The overall architecture of the CNN used in this work, which includes an input layer, multiple alternating convolution and max-pooling layers, and one fully connected classification layer.}
	\label{fig:fig-2}       
\end{figure*}

\subsection{Convolutional neural network }
The CNN structure was first time proposed by Fukushima in 1980 \cite{23}. However, it has not been widely used because the training algorithm was not easy to use. In 1990s, LeCun et al. applied a gradient-based learning algorithm to CNN and obtained successful results \cite{24}. After that, researchers further improved CNN and reported good results in pattern recognition. Recently, Cireşan et al. applied multi-column CNNs to recognize digits, alpha-numerals, traffic signs, and the other object class \cite{26,25}. They reported excellent results and surpassed conventional best records on many benchmark databases, including MNIST \cite{50} handwritten digits database and CIFAR-10 \cite{51}. In addition to the common advantages of DNNs, CNN has some extra properties: it is designed to imitate human visual processing, and it has highly optimized structures to learn the extraction and abstraction of two dimensional (2D) features. In particular, the max-pooling layer of CNN is very effective in absorbing shape variations. Moreover, composed of sparse connection with tied weights, CNN requires significantly fewer parameters than a fully connected network of similar size. Most of all, CNN is trainable with the gradient-based learning algorithm, and suffers less from the diminishing gradient problem. Given that the gradient-based algorithm trains the whole network to minimize an error criterion directly, CNN can produce highly optimized weights. Recently, deep CNN was applied for Hangul handwritten character recognition and achieved the best recognition accuracy \cite{27}. 

Figure \ref{fig:fig-2} shows an overall architecture of CNN that consists with two main parts: feature extraction and classification. In the feature extraction layers, each layer of the network receives the output from its immediate previous layer as its input, and passes the current output as input to the next layer. The CNN architecture is composed with the combination of three types of layers: convolution, max-pooling, and classification. Convolutional layer and max-pooling layer are two types of layers in the low and middle-level of the network. The even numbered layers work for convolution and odd numbered layers work for max-pooling operation. The output nodes of the convolution and max-pooling layers are grouped in to a 2D plane which is called feature mapping. Each plane of the layer usually derived with the combination of one or more planes of the previous layers. The node of the plane is connected to a small region of each connected planes of the previous layer. Each node of the convolution layer extracts features from the input images by convolution operation on the input nodes. The max-pooling layer abstracts features through average or propagating operation on the input nodes.

The higher level features is derived from the propagated feature of the lower level layers. As the features propagate to the highest layer or level, the dimension of the features is reduced depending on the size of the convolutional and max-pooling masks. However, the number of feature mapping usually increased for mapping the extreme suitable features of the input images to achieve better classification accuracy. The outputs of the last feature maps of CNN are used as input to the fully connected network which is called classification layer. In this work, we use the feed-forward neural networks as a classifier in the classification layer, because it has proved better performance compared to some recent works \cite{28,29}. In the classification layer, the desired number of features can be obtained using feature selection techniques depending on the dimension of the weight matrix of the final neural network, then the selected features are set to the classifier to compute confidence of the input images. Based on the highest confidence, the classifier gives outputs for the corresponding classes that the input images belong to. Mathematical details of different layers of CNN are discussed in the following section.

\subsubsection{Convolution layer}
In this layer, the feature maps of the previous layer are convolved with learnable kernels such as (Gaussian or Gabor). The outputs of the kernel go through linear or non-linear activation functions such as (sigmoid, hyperbolic tangent, softmax, rectified linear, and identity functions) to form the output feature maps. In general, it can be mathematically modeled as

\begin{equation}
	x_j^l = f\left(\sum_{i\in M_j} x_i^{l-1} k_{ij}^l+ b_j^l\right)
\end{equation}

\noindent
where $x_j^l$ is the outputs of the current layer, $x_i^{l-1}$ is previous layer outputs, $k_{ij}^l$ is kernel for present layer, and $b_j^l$ is the bias for current layer. $M_j$ represents a selection of input maps. For each output map is given an additive bias $b$. However, the input maps will be convolved with distinct kernels to generate the corresponding output maps. For instant, the output maps of $j$ and $k$ both are summation over the input $i$ which is in particular applied the $j^{th}$ kernel over the input $i$ and takes the summation of its and same operation are being considered for $k^{th}$ kernel as well.

\subsubsection{Subsampling layer}
The subsampling layer performs downsampling operation on the input maps. In this layer, the input and output maps do not change. For example, if there are $N$ input maps, then there will be exactly $N$ output maps. Due to the downsampling operation, the size of the output maps will be reduced depending on the size of the downsampling mask. In this experiment, $2\times2$ downsampling mask is used. This operation can be formulated as

\begin{equation}
	x_j^l = f\left(\beta_j^l down(x_j^{l-1})+ b_j^l \right)
\end{equation}
\noindent
where $down(\cdot)$ represents a subsampling function. This function usually sums up over $n \times n$ block of the maps from the previous layers and selects the average value or the highest values among the $n \times n$ block maps. Accordingly, the output map dimension is reduced to $n$ times with respect to both dimensions of the feature maps. The output maps finally go through linear or non-linear activation functions. 

\subsubsection{Classification layer}
This is a fully connected layer which computes the score for each class of the objects using the extracted features from convolutional layer. In this work, the size of the feature map is considered to be $5\times5$ and a feed-forward neural net is used for classification. As for the activation function, sigmoid function is employed as suggested in most literatures.

\subsubsection{Back-propagation}
In the BP steps in CNNs, the filters are updated during the convolutional operation between the convolutional layer and immediate previous layer on the feature maps and the weight matrix of each layer is calculated accordingly. 

\subsection{CNN with dropout}
The combination of the prediction of different models is a very effective way to reduce test errors \cite{31,32}, but it is computationally expensive for large neural networks that can take several days for training. However, there is a very efficient technique for the combination models named ``dropout" \cite{33}. In this model, the outputs of hidden layer neurons are set to be zero if the probability is less than or equal to a certain value, for example $0.5$. The neurons that are ``dropped out" in the way to forward pass that do not have any impact on BP. Dropout reduces complexity of the network because of co-adaptation of neurons, since one set of neurons are not rely on the presence of another set of neurons. Therefore, it is forced to learn more robust features that are useful in aggregation with many different random subsets of the other neurons. However, one of the drawbacks of the dropout operation is that it may take more iterations to reach the required convergence level. In this work, dropout is applied in the first two fully-connected layers in Fig. \ref{fig:fig-2}. 

\begin{figure}[h]
	\includegraphics[scale=0.33]{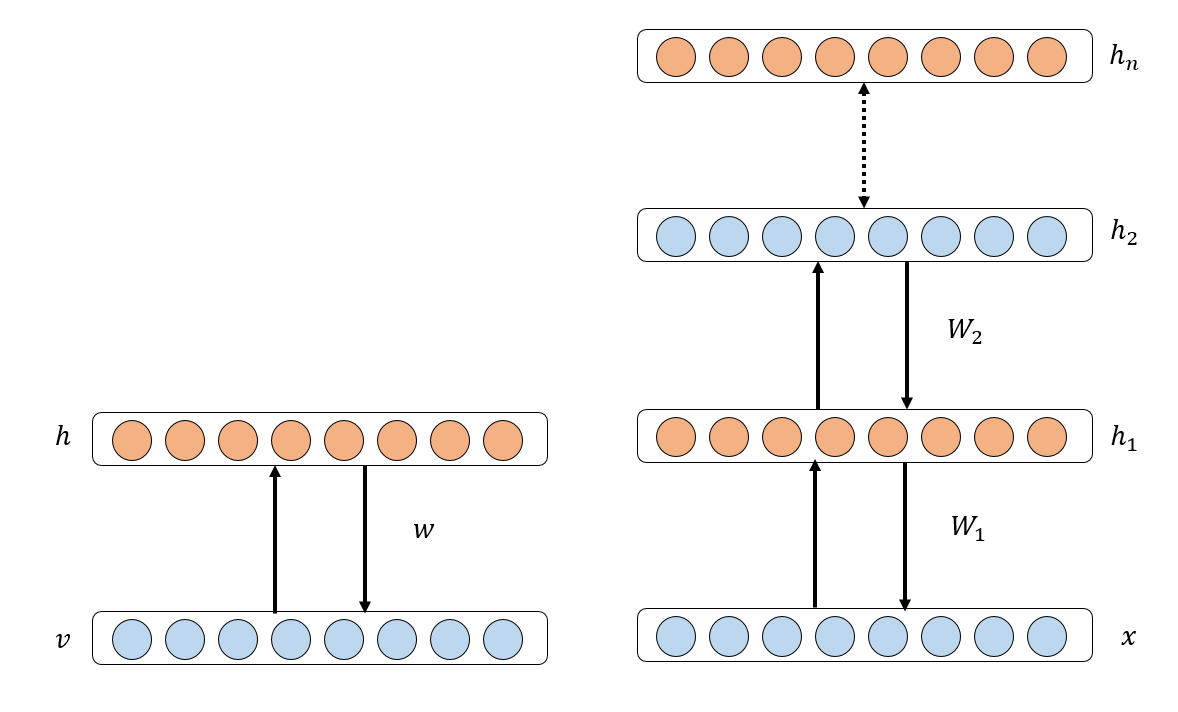}
	\caption{Illustration of RBM (left) and DBN (right).}
	\label{fig:fig-3}       
\end{figure}

\subsection{Restricted Boltzmann Machine (RBM)}
RBM is based on Markov Random Field (MRF) and it has two units: binary stochastic hidden unit and binary stochastic visible unit. It is not mandatory of the unit to be Bernoulli random variable and can in fact have any distribution in the exponential family \cite{34}. Besides, there is connection between hidden to visible and visible to hidden layer but there is no connection between hidden to hidden or visible to visible units. The pictorial representation of RBM is shown in Fig. \ref{fig:fig-3}.

The symmetric weights on the connections and biases of the individual hidden and visible units are calculated based on the probability distribution over the binary state vector of $v$ for the visible units via an energy function. The RBM is an energy-based undirected generative model which uses a layer of hidden variables to model the distribution over visible variable in the visible units \cite{36}. The undirected model of the interactions between the hidden and visible variables of both units is used to confirm that the contribution of the probability term to posterior over the hidden variables \cite{37}. 

Energy-based model means that the likelihood distribution over the variables of interest is defined through an energy function. It can be composed from a set of 
observable variables $V={v_i}$ and a set of hidden variables $H={h_i}$ where $i$ is the node in the visible layer and $j$ is the node in the hidden layer. It is restricted in the sense that there are no visible-visible or hidden-hidden connections. 

The input values correspond to the visible units of RBM for observing their and the generated features correspond to the hidden units. A joint configuration, $(v,h)$ of the visible and hidden units has an energy given by \cite{34}:

\begin{equation}
	E(v,h;\theta) = -\sum_{i} a_i v_i -\sum_{j} b_j h_j - \sum_{i}\sum_{j} v_i  h_j w_{ij}
\end{equation}
\noindent
where $\theta=(w,b,a)$,  $v_i$  and $h_j$ are the binary states of visible unit $i$ and hidden unit $j$. $w_{ij}$ is the symmetric weight in between visible and hidden units, and $a_i$, $b_j$ are their respective biases. The network assigns a probability to every possible pair of a visible and a hidden vector via this energy function as
\begin{equation}
	p(v,h) = \frac{1}{Z} e^{-E(v,h;\theta)}
\end{equation}
\noindent
where the “partition function”, $Z$ is given by summing over all possible pairs of visible and hidden vectors as follows
\begin{equation}
	Z = \sum_{v,h} e^{-E(v,h)}
\end{equation}

The probability which the network assigns to a visible vector $v$, is generated through the summation over all possible hidden vectors as
\begin{equation}
	p(v) = \frac{1}{Z} \sum_{h} e^{-E(v,h;\theta)}
\end{equation}

The probability for training inputs can be improved by adjusting the symmetric weights and biases to decrease the energy of that image and to increase the energy of other images, especially those have low energies, and as a result, it makes a huge contribution for partitioning function. The derivative of the $log$ probability of a training vector with respect to symmetric weight is computed as
\begin{equation}
	\frac{\partial \log p(v)}{\partial w_{ij}} = \langle v_j h_j \rangle_{d} - \langle v_j h_j \rangle_{m}
\end{equation}
\noindent
where $ \langle \cdot \rangle_d $ represents the expectations for the data distribution and $ \langle \cdot \rangle_m $ denotes the expectations under the model distribution. It contributes to a simple learning rule for performing stochastic steepest ascent in the $log$ probability on the training data:

\begin{equation}
	w_{ij} = \epsilon \frac{\partial \log p(v)}{\partial w_{ij}}
\end{equation}
\noindent
where $\epsilon$ is the learning rate. Due to no direct connectivity between hidden units in an RBM, it is easy to get an unbiased sample of $\langle v_j h_j \rangle_{d}$. Given a randomly selected training image $v$, the binary state $h_j$ of each hidden unit $j$ is set to $1$ with probability
\begin{equation}
	p(h_j = 1|v) = \sigma\left(b_j + \sum_{i} v_i w_{ij} \right)
\end{equation}
\noindent
where $\sigma(\cdot)$ is the logistic sigmoid function. Similarly, because there is no direct connections between visible units in RBM, it is easy to compute an unbiased sample of the state of a visible unit, given a hidden unit
\begin{equation}
	p(v_i=1|h) = \sigma\left(a_i + \sum_{j} h_j w_{ij} \right)
\end{equation}

However, it is much more difficult to generate unbiased sample of $\langle v_j h_j \rangle_{m}$. It can be done in the beginning at any random state of visible layer and performing alternative Gibbs sampling for very long period of time.  Gibbs sampling consists of updating all of the hidden units in parallel using Eq. (9) in one alternating iteration followed by updating all of the visible units in parallel using Eq. (10). 

However, a much faster learning procedure has been proposed by Hinton \cite{35}. In this approach, it starts by setting of the states of the visible units to a training vector. Then the binary states of the hidden units are all computed in parallel according to Eq. (9). Once binary states are selected for the hidden units, a ``reconstruction" is generated by setting each $v_i$ to 1 with a probability given by Eq. (10). The change in a weight matrix can be written as

\begin{equation}
	\triangle w_{ij} = \epsilon\left(\langle v_j h_j \rangle_{d} - \langle v_j h_j \rangle_{r} \right)
\end{equation}
where $ \langle \cdot \rangle_r $ represents the expectations for the model distribution from the ``reconstruction" states.

A simplified version of the same learning rule that uses for the states of individual units. However,  the pairwise products approach is used for the biases. The learning rule closely approximates the gradient of another objective function called the Constrictive Divergence (CD) \cite{36} which is different from Kullback-Liebler divergence. However, it work well to achieve better accuracy in many applications. CD$_n$ is used to represent learning using $n$ full steps of alternating Gibbs sampling.

The pre-training procedure of RBM of a DBN can be utilized to initialize the weight of DNNs, which can be discriminatively fine-tuned by BP error derivative. There are different activation functions have been used such as sigmoid \cite{38}, hyperbolic tangent \cite{38}, softmax \cite{39}, and rectified linear \cite{29} in different implementations using DBN. In this work, a sigmoid function is considered.

\begin{figure*}
	\centering
	\includegraphics[scale=0.5]{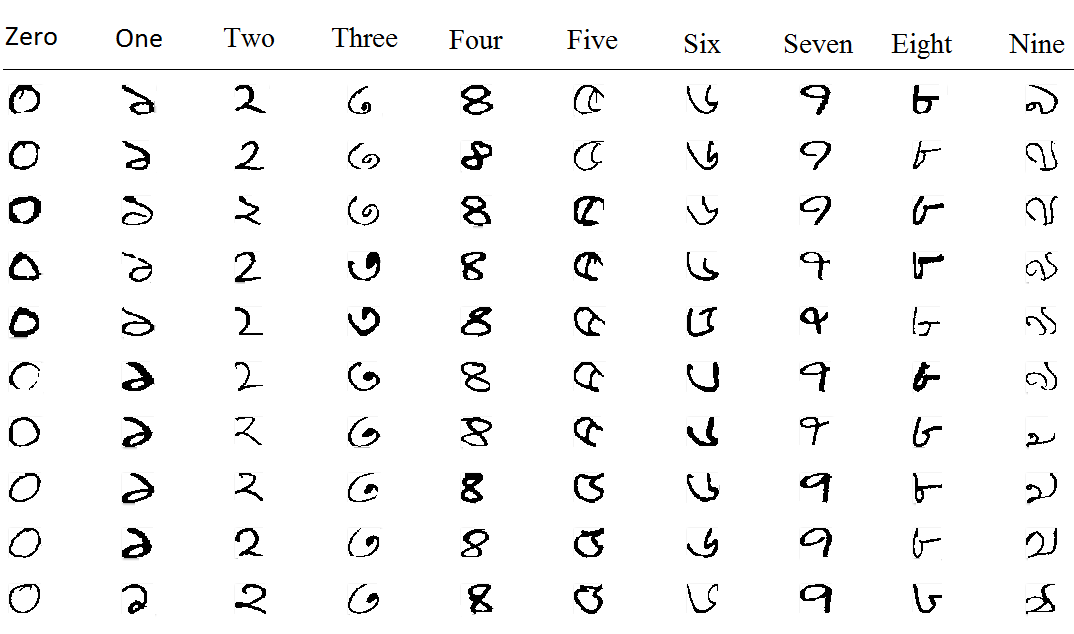}
	\caption{Sample handwritten Bangla numeral images: row 1 indicates the actual digit class and rows 2-11 illustrate some randomly selected handwritten Bangla numeral images.}
	\label{fig:fig-4}       
\end{figure*}

\subsection{Deep belief network}
A hidden unit of every layer learns to represent the feature perfectly that is determined by the higher order correlation in the original input data as shown in Fig. \ref{fig:fig-3}. The main idea behind the training concept of a DBN is to train a sequence of RBMs with the model parameter $\theta$. The trained RBM generates the probability of an output vector for the visible layer, $p(v|h,\theta)$ in conjunction with the hidden layer distribution, $p(h,\theta)$, so the probability of generating a visible layer output as a vector $v$, can be written as:
\begin{equation}
	p(v) = \sum_{h} p(h,\theta) p(v|h,\theta)
\end{equation}

After learning the parameters θ and  $p(v|h,\theta)$ is kept while $p(h,\theta)$ can be replaced by an improved model that is learned by treating the hidden activity vectors $H=h$ as the training data (visible layer) for another RBM. This replacement improves a variation lower bound on the probability of the training data under the composite model \cite{28}. The following three rules can be resulting in the study of according to \cite{41}:

\begin{itemize}
	\renewcommand{\labelitemi}{$\circ$}
	\item If the number of hidden units in the top level of the network crosses a predefined threshold; the performance of DBN essentially flattens at around certain accuracy.
	\item The trend of the performance decreases as the number of layers increases.
	\item The performance of RBMs upgrades during training as the number of iteration increases.
\end{itemize}

DBNs can be used as a feature extraction method for dimensionality reduction where the class labels is not required with BP in the DBN architecture (unsupervised training) \cite{42}. On the other hand, when the associated labels of the class is incorporated with feature vectors, DBNs is used as a classifier. There are two general types of classifiers depending on architecture which are BP-DBNs and Associate Memory DBNs (AM-DBN) \cite{33}. When the number of the possible class is very large, then the distribution of the frequencies for different classes is far from uniform for both architectures. However, it may sometimes be advantageous to use a different encoding for the class targets than the standard one-of-$K$ softmax encoding \cite{34}. In our proposed method, DBNs is used as a classifier.

\begin{table*}
	
	\caption{Parameters setup for CNN}
	\begin{tabular}{llllll}
		\hline\noalign{\smallskip}
		Layer & Operation of Layer & Number of feature maps & Size of feature maps & Size of window & Number of parameters \\
		\noalign{\smallskip}\hline\noalign{\smallskip}
		$C_1$ & Convolution & 32 & $28\times28$ & $5\times5$ & 832\\
		$S_1$ & Max-pooling & 32 & $14\times14$ & $2\times2$ & 0 \\
		$C_2$ & Convolution & 64 & $10\times10$ & $5\times5$ & 53,248 \\
		$S_2$ & Max-pooling & 64 & $5\times5$ & $2\times2$ & 0 \\
		$F_1$ & Fully connected & 312 & $1\times1$ & N/A & 519,168 \\
		$F_2$ & Fully connected & 10 & $1\times1$ & N/A & 3,130 \\
		\noalign{\smallskip}\hline
	\end{tabular}
\end{table*}

\begin{figure*}
	\centering
	\includegraphics[scale=0.1]{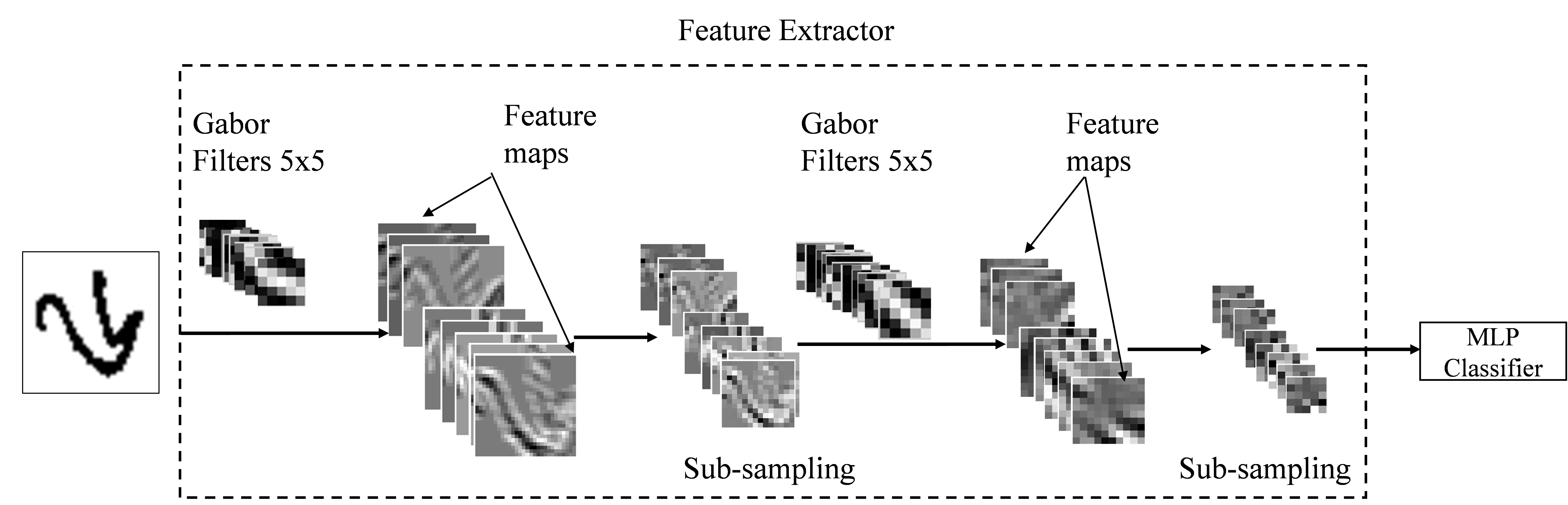}
	\caption{Visualization of feature extraction in CNN.}
	\label{fig:fig-5}       
\end{figure*}

In this paper, we employ and evaluate the power of DNNs including DBN, CNN and CNN with dropout on HBDR. We also test the performance of CNN with random filters, CNN with dropout, CNN with dropout and initial random filters, and CNN with dropout and Gabor features. Finally, experimental results and performance evaluation against SVM are provided.

\section{Experimental results and discussion}
\subsection{Dataset description}
We evaluated the performance of DBN and CNN on a benchmark dataset called CMATERdb $3.1.1$ \cite{45,17}. This dataset contains 6000 images of unconstrained handwritten isolated Bangla numerals. Each digit has 600 images of $32\times32$ pixels. Some sample images of the database are shown in Fig. \ref{fig:fig-4}. There is no visible noise can be seen in visual inspection. However, variability in writing style due to user dependency is quite high. The data set was split into a training set and a test set. We randomly selected 5000 images (500 randomly selected images of each digit) for the training set and the test set contains the remaining 1000 images. 

\subsection{CNN structure and parameters setup}
In this experiment, we used six layers of convolutional neural networks. Two layers for convolution, two layers for subsampling or pooling, and final one layer for classification. The first convolution layer has 32 output mapping and the second one has 64 output mapping. The parameter of convolutional network is calculated according to the following manner: $32\times32$ image is taken as input. The output of the convolutional layer is $28\times28$ with $32$ feature maps. The size of the filter mask is $5\times5$ for the both convolution layers. The number of parameters are used to learn is $(5\times5+1) \times32=832$ and the total number of connection is $28\times28\times(5\times5+1) \times32=652,288$. For the first subsampling layer, the number of trainable parameters is $0$ and the size of the outputs of subsampling layer is $14\times14$ with $32$ feature maps. According to this way the remaining two convolutional and subsampling layers' parameters are calculated. The learning parameters for second convolution layer is $((5\times 5+1)\times32)\times64=53,248$ and $0$ for convolutional and sub-sampling layers, respectively. In the fully connected layer, number of feature maps is an empirically chosen number which is $312$ from the previous max-pooling layer provides outputs with $64$ maps and $5\times5$ size of output for each input. The number of parameters for the first fully connected layer is: $312\times64\times(5\times5+1) =519,168$, whereas the amount of the final layer's parameter is: $10\times(312+1) =3,130$. Total number of parameters is 576,378. All the parameters with respect to the corresponding layers is stated in Table 1, and Fig. \ref{fig:fig-5} illustrates a corresponding feature extraction process in CNN. 

\begin{figure}[h]
	\centering
	\includegraphics[scale=0.29]{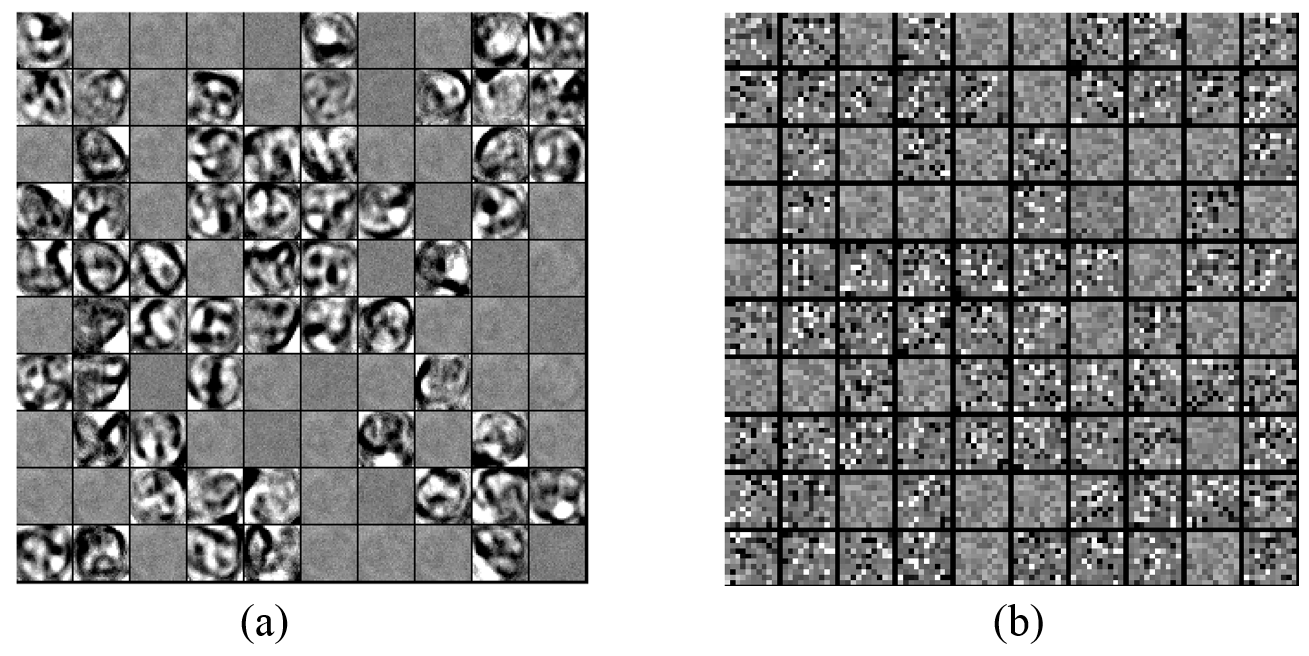}
	\caption{Learned weights of (a) layer 1 and (b) layer 2 in DBN.}
	\label{fig:fig-6}       
\end{figure}

\subsection{DBN structure and parameters setup} 
\begin{figure}[h]
	\centering
	\includegraphics[scale=1]{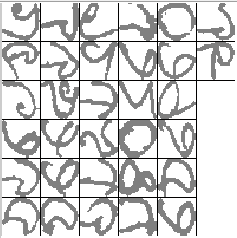}
	\caption{Misclassified digits by DBN.}
	\label{fig:fig-7}       
\end{figure}

\begin{figure}[h]
	\centering
	\includegraphics[scale=0.25]{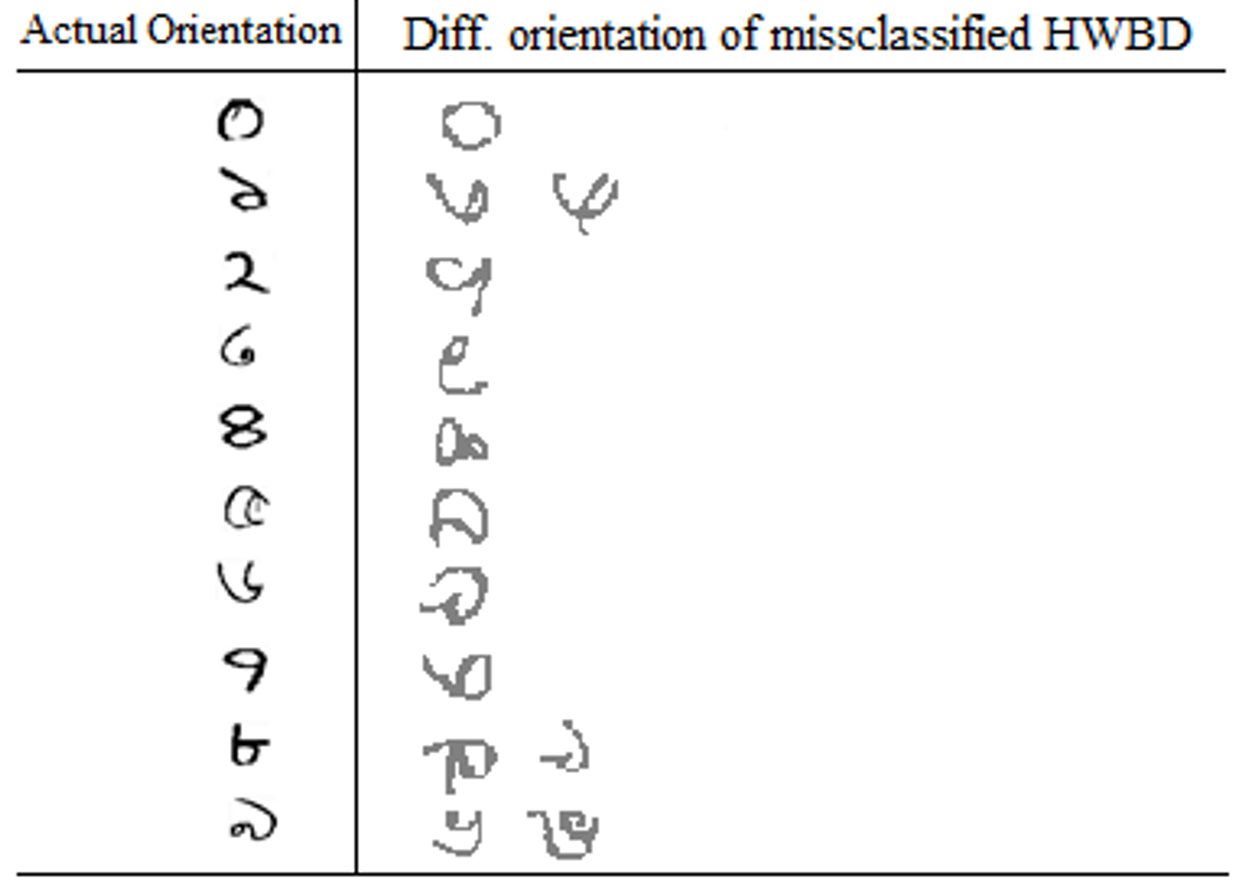}
	\caption{Orientation of actual and misclassified digits in the database.}
	\label{fig:fig-8}       
\end{figure}

In this experiment, a DBN with two RBM based hidden layers trained with Bernoulli hidden and visible units has been implemented. The soft-max layer is used as final prediction layer in DBN. In the hidden layer, 100 hidden units have been considered with learning rate 0.1, momentum 0.5, penalty $2\times e^{-4}$ and batch size 50. Contractive Divergence, which is an approximate Maximum Likelihood (ML) learning method, has been considered in this implementation. The learned weights for the respective hidden layers of DBN are shown in Fig. \ref{fig:fig-6}. Misclassified Bangla handwritten digits using DBN technique are shown in Fig. \ref{fig:fig-7}. From the misclassified image, it can be clearly observed that the digits which are not recognized accurately are written in different orientations. Fig. \ref{fig:fig-8} shows some examples of Handwritten Bangla Digit (HWBD) with actual orientation and the orientation of digits in the database that are recognized incorrectly by DBN.

\subsection{Performance evaluation}
\begin{figure}[h]
	\centering
	\includegraphics[scale=0.6]{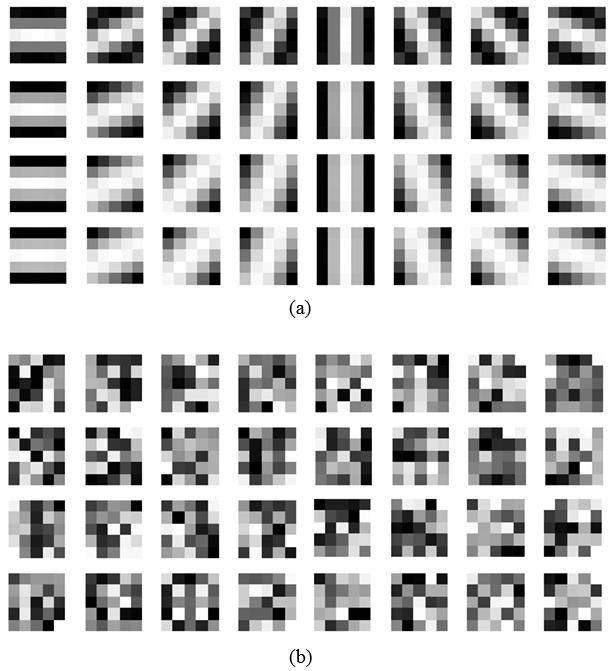}
	\caption{Examples of (a) Gabor filters and (b) Gaussian filters.}
	\label{fig:fig-9}       
\end{figure}

\begin{figure}[h]
	\centering
	\includegraphics[scale=0.55]{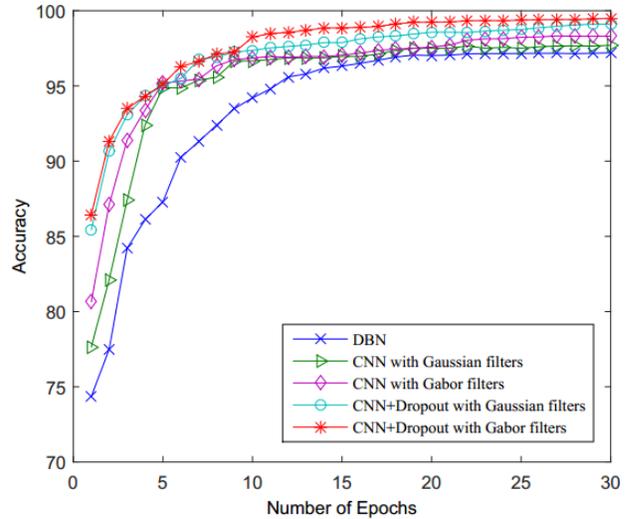}
	\caption{Comparison of testing accuracy for 30 iterations.}
	\label{fig:fig-10}       
\end{figure}

The experimental results and comparison of different approaches are shown in Table 2. There are thirty iterations have been considered in for training and testing in this experiment. The testing accuracy is reported. SVM provides $95.5\%$ testing accuracy, whereas DBN produces $97.20\%$. Besides, CNN with random Gaussian filter provides accuracy of $97.70\%$, while CNN with Gabor kernels provides around $98.30\%$ which is higher than standard CNN with Gaussian filters. Fig. \ref{fig:fig-9} shows examples of the Gabor ($5\times5$) and Gaussian kernels ($5\times5$) used in the experiment. On the other hand, the dropout based CNN with Gaussian and Gabor filters provide $98.64\%$ and $98.78\%$ testing accuracy for HBDR, respectively. It is observed that the CNN with dropout and Gabor filter outperforms CNN with dropout and random Gaussian filter. Thus, it can be concluded that Gabor feature in CNN is more effective for HBDR. According to the Table 2, it is also clear that the CNN with dropout and Gabor filter gives the best accuracy compared to the other most influential machine learning methods such as SVM, DBN, and standard CNN. Fig. \ref{fig:fig-10} shows the recognition performance of DBN, CNN, CNN with dropout, Gaussian filters and Gabor filters for 30 iterations. This figure illustrates the minimum number of iterations required for achieving the best recognition accuracy. In this case, it can be seen that after around fifteen iteration we have reached almost the maximum accuracy. 

\begin{table}[t]
	\caption{Comparison of recognition performance (Bold font indicates the highest accuracy)}
	\label{sample-table1}
	\vskip 0.15in
	\begin{center}
		\begin{small}
			\begin{sc}
				\begin{tabular}{lcccr}
					\hline
					\abovespace
					Methods & Accuracy \\ 
					\hline
					\abovespace
					SVM  & 95.50\% \\
					DBN  & 97.20\%\\
					CNN + Gaussian  & 97.70\% \\
					CNN + Gabor  & 98.30\% \\
					CNN + Gaussian + Dropout  & 98.64\%\\
					CNN + Gabor + Dropout  & \bf{98.78\%}\\ 
					\hline
				\end{tabular}
			\end{sc}
		\end{small}
	\end{center}
	\vskip -0.1in
\end{table}



\begin{table*}
	\caption{Comparison with state-of-the-arts (Bold font indicates the highest accuracy in each case of training and testing samples)}
	\label{sample-table2}
	\vskip 0.15in
	\begin{center}
		\begin{small}
			\begin{sc}
				\begin{tabular}{lcccr}
					\hline
					\abovespace
					Methods & Training / Testing Samples & Accuracy  \\
					\hline
					\abovespace
					MLP \cite{16} & 4000 / 2000 & 96.67\%  \\
					MPCA+QTLR \cite{45}  & 4000 / 2000 & 98.55\%  \\
					GA \cite{17}  & 4000 / 2000 & 97.00\%  \\
					SRC \cite{14} & 5000 / 1000 & 94.00\%  \\  
					\hline 
					\abovespace
					Proposed & 4000 / 2000 & \bf98.64\% \\
					         & 5000 / 1000 & \bf98.78\% \\
					\hline
				\end{tabular}
			\end{sc}
		\end{small}
	\end{center}
	\vskip -0.1in
\end{table*}

\subsection{Comparison with the state-of-the-arts}

Lastly, we also compare our proposed DL method (CNN + Gabor + Dropout) with the state-of-the-art techniques, such as MLP \cite{16}, Modular Principal Component Analysis (MPCA) with Quad Tree based Longest-Run (MPCA+QTLR) \cite{45}, Genetic Algorithm (GA) \cite{17}, Simulated Annealing (SA) \cite{17}, and Sparse Representation Classifier (SRC) \cite{14} based algorithms for HBDR on the same database. The recognition performance of those approaches is listed in Table 3. As shown in this table, the number of training and testing samples are varying with respect to the methods. Thus, for fair comparison, we conducted another experiments using 4000 training and 2000 testing samples, and we reached $98.78\%$ accuracy at the $16^{th}$ iteration where it already exceeds all other alternative techniques for HBDR.

\section{Conclusion}
In this research, we proposed to use deep learning approaches for handwritten Bangla digit recognition(HBDR). We evaluated the performance of CNN and DBN with combination of dropout and different filters on a standard benchmark dataset: CMATERdb $3.1.1$. From experimental results, it is observed that CNN with Gabor feature and dropout yields the best accuracy for HBDR compared to the alternative state-of-the-art techniques. Research work is currently progressing to develop more sophisticated deep neural networks with combination of State Preserving Extreme Learning Machine \cite{44} for handwritten Bangla numeral and character recognition.


\bibliographystyle{icml2017}
\bibliography{myReferences}


\end{document}